\documentclass[letterpaper, 10 pt, conference]{ieeeconf}  
\IEEEoverridecommandlockouts                              
\usepackage[utf8]{inputenc}
\usepackage[T1]{fontenc}
\usepackage{hyperref}
\usepackage{graphicx}
\usepackage{float}

\title{\LARGE \bf
Medical Applications of Graph Convolutional Networks Using Electronic Health Records: A Survey
}

\author{%
Garrik Hoyt$^{1}$, 
Noyonica Chatterjee$^{2}$,
Fortunato Battaglia$^{3}$, 
Paramita Basu$^{4}$
\thanks{$^{1}$Department of Computer Science and Engineering, Lehigh University, Bethlehem, PA 18015, USA.}%
\thanks{$^{2}$Plaksha University, Sahibsada Ajit Singh Nagar, Punjab 140306, IND}%
\thanks{$^{3}$Department of Medical Science and Neurology, Hackensack Meridian School of Medicine, Nutley, NJ 07110, USA.}%
\thanks{$^{4}$Pre-Clinical Sciences Department, New York College of Podiatric Medicine at Touro University, New York, NY 10035, USA.}%
}

\begin{document}
\maketitle
\thispagestyle{empty}
\pagestyle{empty}

\begin{abstract}

\par Graph Convolutional Networks (GCNs) have
emerged as a promising approach to machine learning on
Electronic Health Records (EHRs). By constructing a graph
representation of patient data and performing convolutions
on neighborhoods of nodes, GCNs can capture complex relationships and extract meaningful insights to support medical
decision making. This survey provides an overview of the
current research in applying GCNs to EHR data.
We identify the key medical domains and prediction tasks
where these models are being utilized, common benchmark
datasets, and architectural patterns to provide a comprehensive survey of this field. While this is a nascent area of research, GCNs
demonstrate strong potential to leverage the complex information hidden in EHRs. Challenges and opportunities for future work are also discussed.

\end{abstract}


\section{Introduction}

\subsection{Electronic Health Records}

\par Electronic Health Records (EHRs) are digital representations of patients' paper charts. EHRs store a wide range of data such as demographics, medical history, medications and laboratory test results. EHR systems are intended to support healthcare activities directly or indirectly through various interfaces, including evidence-based decision support, quality management, and outcomes reporting \cite{Shickel2018}.

\par EHR data is a valuable resource for advancing medical science and healthcare. In research, it can be used with machine learning and data science methods to derive insights from these often large datasets. These insights can lead to new breakthroughs in medical science. One of the challenges with EHR data is that it is heterogeneous, meaning it has multiple types of data. Heterogeneous data present significant challenges for machine learning models; a model will have to make sense of the various data sources, similar to when a doctor reviews a patient's chart. Additionally, privacy and security of patient data is crucial, often making it difficult to obtain large datasets for training deep learning models.

\subsection{Graph Neural Networks}

\par Graph Neural Networks (GNNs) are a class of deep learning models designed to perform inference tasks on graph-structured data. GNNs have gained attention due to their ability to learn representations that capture the dependencies between nodes in a graph, making them suitable for various applications such as social network analysis, recommendation systems, and molecular property prediction \cite{Wu2021}. For any graph-based model, a graph must first be constructed that captures entities as vertices and models the relationships between them as edges. Edges can be weighted or unweighted, in addition to being directed or undirected depending on the nature of the connections.

\par The concept of GNNs was first introduced by Gori et al. and further developed by Scarselli et al. \cite{Gori2005}, \cite{Scarselli2009}. Early works proposed a recursive neural network architecture to update node representations based on the information from their neighbors until a stable fixed point is reached. In recent years, there has been a surge of interest in GNNs, leading to the development of various GNN architectures, such as Graph Convolutional Networks (GCNs)  Graph Attention Networks (GATs) and Graph Recurrent Networks \cite{Kipf2017}, \cite{Velickovic2018}, \cite{Li2016}.

\subsection{Graph Convolutional Networks}

\begin{figure*}[h]
\centering
\includegraphics[width=0.85\textwidth]{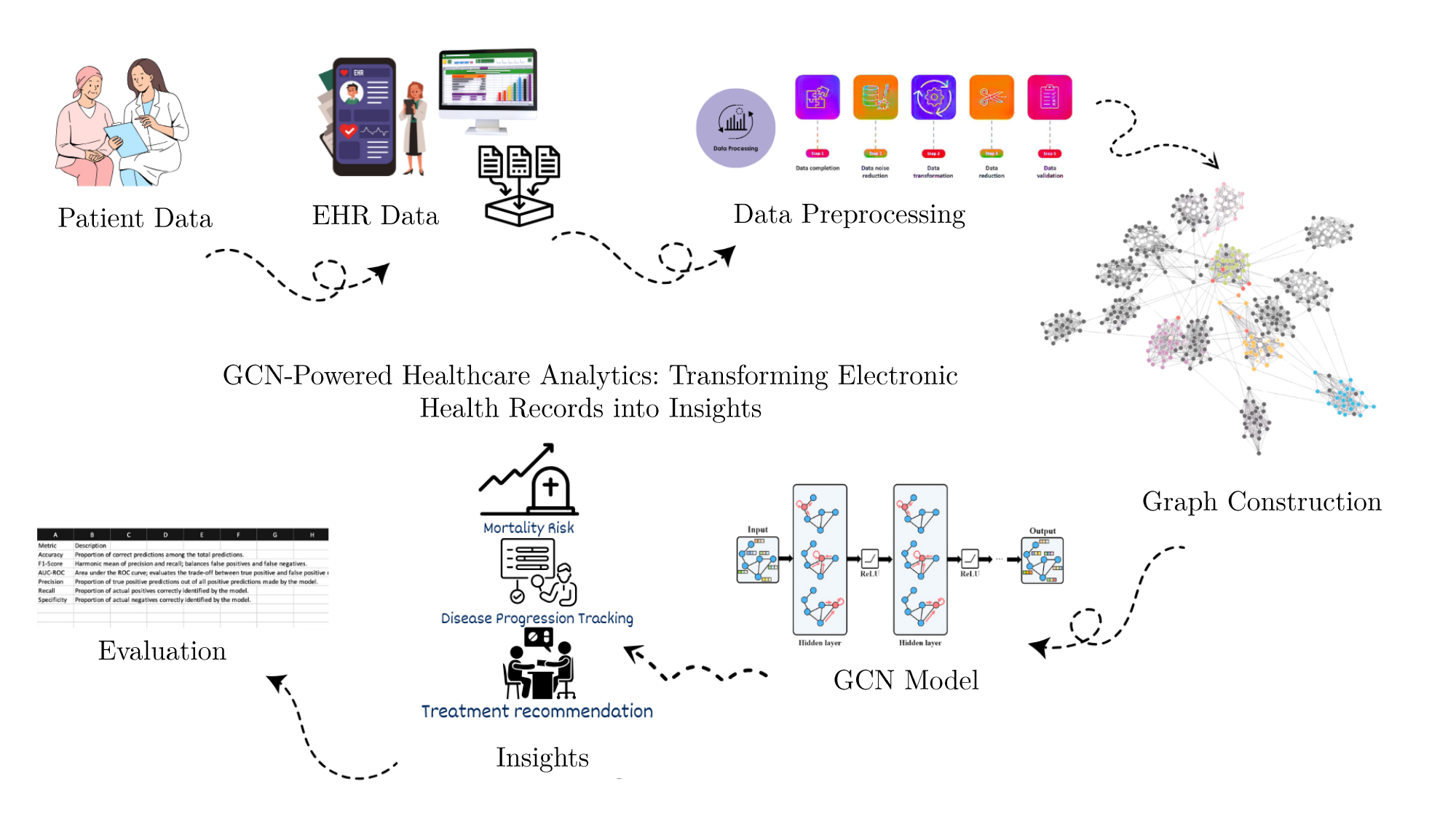}
\caption{Process diagram of how GCNs are used to extract insight from patient data. First, a graph representation of the data is constructed from the EHRs. Next, a GCN performs convolution operations to derive information from the relationships between neighboring nodes in the graph. The swirling in the representation of the GCN conveys the aggregation process performed on neighborhoods in graphs. Results can be used to make informed decisions and gain insight into potential causal factors.}
\label{fig:figure1}
\end{figure*}

\par Graph Convolutional Networks (GCNs) are machine learning models that perform convolution operations on graph-structured data. Graph convolutions were first introduced by Bruna et al. and later extended by Defferrard et al.  using Chebyshev polynomials \cite{Bruna2013}, \cite{Defferrard2016}. Kipf and Welling demonstrated the effectiveness of GCNs in semi-supervised node classification tasks\cite{Defferrard2016}, \cite{Kipf2017}. GCNs have gained attention due to their ability to directly operate on graphs and learn node representations by aggregating information from neighboring nodes \cite{Wu2021}.

\par The key idea behind a GCN is to learn a function that aggregates feature information from a node's neighborhood and generates a new representation for each node. In spectral-based GCNs, the convolution operation is defined in the Fourier domain by computing the eigen decomposition of the graph Laplacian matrix \cite{Bruna2013, Defferrard2016}. However, this approach is computationally expensive and requires the entire graph to be processed simultaneously. Spatial-based GCNs perform the convolution operation directly on the graph by aggregating information from neighboring nodes \cite{Kipf2017, Wu2021}. The general framework of spatial-based GCNs involves a message passing scheme, where each node updates its representation by aggregating the representations of its neighbors \cite{Gilmer2017}. This allows GCNs to capture both the structural information of the graph and the feature information of the nodes, making them suitable for various applications such as node classification, link prediction, and graph classification \cite{Wu2021}.

\par GCN models present an interesting approach to making predictions from EHR data, see Fig. 1 for the generalized process. A graph is first constructed to capture relationships within the patient EHR data. The convolution mechanism uses the properties of neighboring nodes in the graph to extract new representations of the data. When applied to EHR data, a GCN can make decisions using interesting relationships in the data that may be otherwise difficult to capture with other deep learning models. Discovering meaningful connections in EHR data can provide meaningful insights that lead to breakthroughs in medical science.

\par This review presents a focused look at how GCNs are being used with EHR data. While there are some published articles on this emerging topic, there are currently no papers surveying the literature in detail. The contributions of this survey are: a summary of commonly used architectures, medical fields  and datasets. This survey paper can provide a potential starting point for future research. The overview detailed in this article can be used to develop a more detailed review of effective methodologies and applications of GCNs with EHR data as more research is published.

\section{Methods}

\subsection{Search Strategy and Sources}

\par A thorough review of papers covering GCNs trained on EHR data in different medical fields was conducted. The PRISMA protocol (Preferred Reporting Items for Systematic Reviews and Meta-Analyses) formalized the selection process, although this paper is not a systematic review or meta-analysis \cite{Page2021}. Lehigh University's ASA Library Catalog was queried for papers for this review. This database search was performed in April 2024. Additional references from a survey paper were included \cite{OssBoll2024}. These papers were categorized as "other", since the database source since they did not come as part of the initial search. Papers from the following databases were included in the results: MEDLINE Ultimate, Academic Search Ultimate, Other, Business Source Complete, Supplemental Index and Complementary Index. The query used to retrieve papers was: ("graph convolutional" OR "GCN") AND (''Electronic health record'' OR ''EHR'' OR ''Electronic medical record'' OR ''EMR'' OR ''electronic health data'). This query was limited to peer-reviewed papers written in English. The publication date range of the query was not limited, since the results were all within the past five years.

\subsection{Eligibility Criteria}

\par This review focused on papers that applied GCN models to different medical domains, using EHRs as a data source. The exclusion criteria used for the review process included: no GCN, no EHR data, no full-text access, and out of scope. Here, out of scope refers to a paper that did not focus on performing a specific medical or healthcare-centric task. Abstracts were reviewed prior to full-text assessment, where papers were categorize according to our organization schema.

\subsection{Data Extraction, Synthesis, and Analysis}

\par Papers were categorized by medical field according to the specified aims of the authors, as well as the nature of the tasks the proposed model performed. We recorded whether papers used only a GCN in their proposed solution, or a hybrid approach using multiple different model architectures. The additional models used in papers proposing a hybrid solution were categorized by model type. Data sources for each paper were documented and summarized. Review and categorization data were stored in a spreadsheet before being loaded into R Studio version 2022.12.0 to summarize and visualize the data.

\begin{figure*}[h!]
\centering
\includegraphics[width = .75\textwidth]{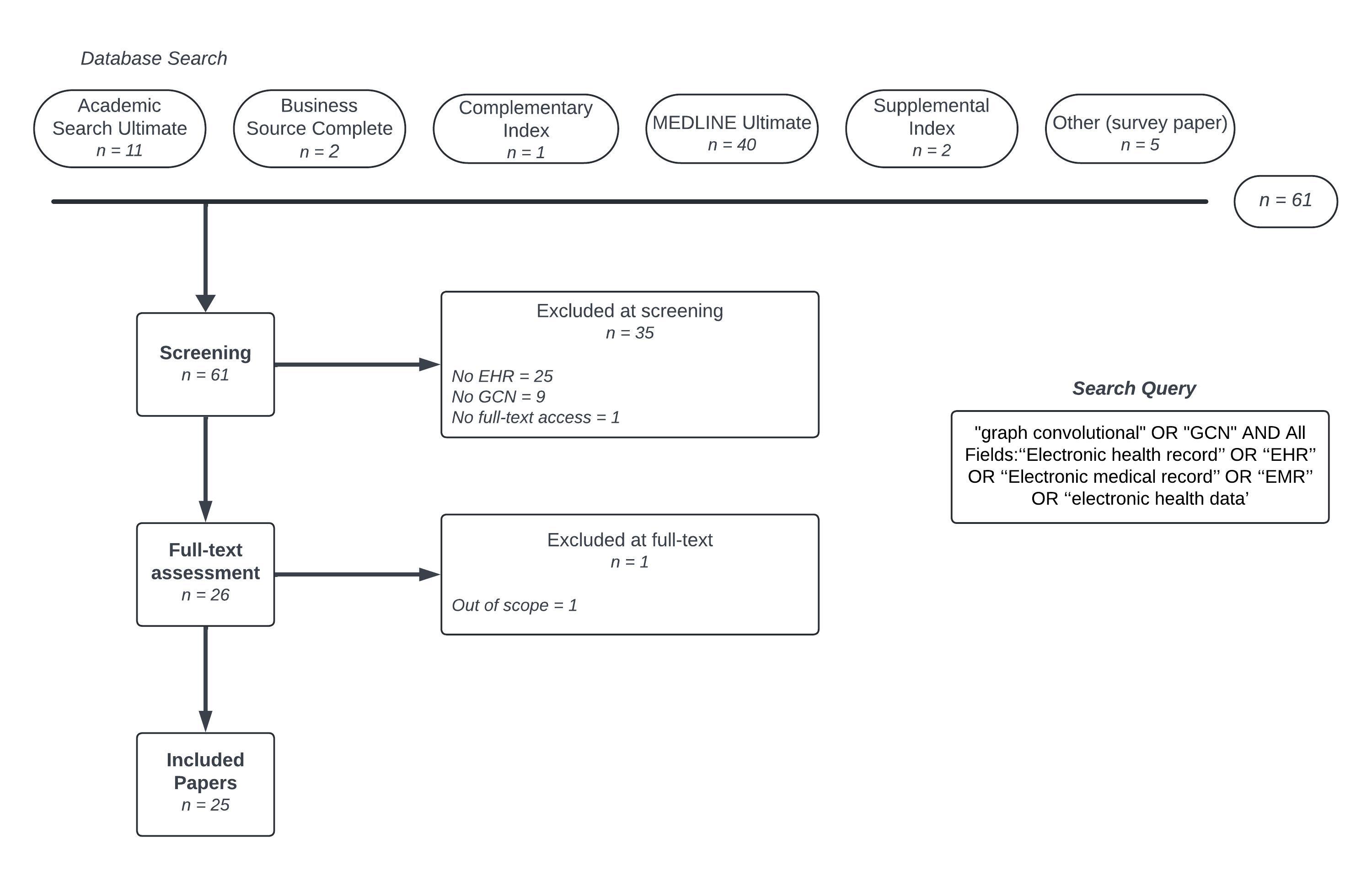}
\caption{PRISMA flowchart for the literature search and selection process for the review.}
\label{fig:figure2}
\end{figure*}

\subsection{Summary of Study Selection}

\par The library catalog query resulted in 56 papers. Five additional papers were included after being found in a survey paper \cite{OssBoll2024, ZhangDiagnosis2024, Shi2021, Wang2021, Wang2020, Yuan2020}. We performed a full-text review of 25 papers, after excluding 35 papers. One paper was excluded for being out of scope of this survey, as it focused more on blockchain and security \cite{Wang2020a}. The full process is outlined as a PRISMA diagram in Fig. 2. An overview of the main findings from reviewing the 25 papers is presented Table 1.

\section{Results}

\par In the Results section, we begin by presenting a high-level overview of the key outcomes of this review. We then summarize the various medical fields that we found in the literature review. We also report how many articles employ a hybrid solution and which models were used in these solutions. Lastly, we identify common EHR data sources across the selected papers.

\begin{figure}[b!]
\centering
\includegraphics[width=0.8\columnwidth]{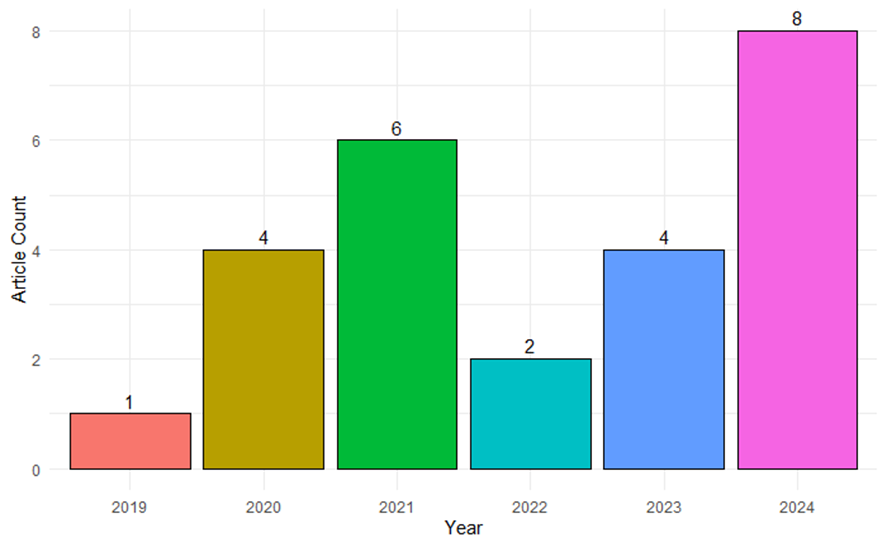}
\caption{Distribution of included articles by year. The cutoff date for selection was April 27, 2024.}
\label{fig:figure3}
\end{figure}

\subsection{Overview of Study Characteristics}

\par We can see in Fig. 3 that using GCNs with EHR data to perform medical tasks is a recently developing trend. Although there are not many published papers on this subject, in only January through April of 2024, we already see eight articles published that were selected for this review. Table 1 shows that hybrid models are commonly applied to problems in this domain. Many studies are using publicly available benchmark datasets, such as the MIMIC-III and MIMIC-IV \cite{Johnson2016, Johnson2023}.

\begin{table*}[h!]
\begin{tabular}{llp{0.2\textwidth}llp{0.3\textwidth}}
\hline
Ref & Year & Dataset used                                                             & Medical Area                & Hybrid Models & Additional Model(s)      \\ \hline
24  & 2019 & 2010 i2b2/VA                                                             & Medical Informatics & X             & LSTM                     \\
21  & 2020 & MIMIC-III                                                                & Medical Informatics & X             & CNN                      \\
23  & 2020 & MIMIC-III; UTP                                                           & Medical Informatics & X             & LSTM                     \\
16  & 2020 & MIMIC-III                                                                & Critical Care               & X             & RNN                      \\
17  & 2020 & Real-world data; MIMIC-III                                               & Medical Informatics & X             & Mutual Attentive Network \\
40  & 2021 & Flatiron Health and Foundation Medicine NSCLC clinico-genomic database   & Oncology                    & X             & MGAE                     \\
31  & 2021 & Shanghai Hospital Development Center Clinical Indicator Terminology Base & Medical Informatics & X             & BERT                     \\
27  & 2021 & MedSTS                                                                   & Medical Informatics & X             & BERT                     \\
14  & 2021 & MIMIC-III                                                                & Medical Informatics &               & -                        \\
15  & 2021 & CHECK                                                                    & Medical Informatics &               & -                        \\
35  & 2021 & Real-world data                                                          & Hematology                  &               & -                        \\
39  & 2022 & COEMRs; C-EMRs                                                           & Obstetrics                  & X             & BERT                     \\
30  & 2022 & NMEDW; MIMIC-III                                                         & Medical Informatics &               & -                        \\
26  & 2023 & Real-world data; CCKS-2019                                               & Medical Informatics & X             & BERT, CRF                \\
38  & 2023 & Real-world data                                                          & Diabetic Neuropathy         &               & -                        \\
22  & 2023 & MIMIC-III; MIMIC-IV                                                      & Medical Informatics & X             & RNN                      \\
33  & 2023 & MIMIC-III                                                                & Critical Care               &               & -                        \\
32  & 2024 & eICU-CRD                                                                 & Critical Care               & X             & BERT                     \\
25  & 2024 & 20NG, R8, R52, OHSUMED, MR, MedLit                                       & Medical Informatics & X             & GloVe, BiRNN             \\
36  & 2024 & Real-world data                                                          & Pulmonology                 &               & -                        \\
28  & 2024 & IgA Nephropathy Dataset; MIMIC-III                                       & Medical Informatics & X             & MLP                      \\
37  & 2024 & MIMIC-III                                                                & Cardiology                  & X             & BERT                     \\
13  & 2024 & Real-world data                                                          & Medical Informatics &               & -                        \\
29  & 2024 & Real-world data                                                          & Medical Informatics &               & -                        \\
34  & 2024 & MIMIC-IV + Mayo Clinic data                                              & Hematology                  &               & -                       
\end{tabular}
\caption{A summary of GCN use across multiple studies using EHR data.}
\label{tab:Table 1}
\end{table*}

\subsection{Medical Fields}

\par Articles that were not focused on a specific medical field were categorized as General Medical Informatics, which broadly encompasses the use of technology to improve healthcare. Many models in this category focused on diagnosis prediction or code classification \cite{Wu2024, ZhangDiagnosis2024, Li2023, Shi2021, Wang2021, Lee2020, Yuan2020, Li2019}. Two of these models performed text processing tasks \cite{Wang2024, Li2023a}. Other tasks performed in this field included similar patient retrieval, medicine recommendation, terminology base enrichment and relation classification \cite{Gu2024, Zhu2024, Mao2022, Zhang2021, Li2019}.

\par Models applied to the field of critical care performed mortality, readmission, and decompensation prediction \cite{Sun2024, LeBaher2023}. Two studies in Hematology used GCNs to perform clinical risk assessment and relation classification \cite{Tariq2024, Mitra2021}. The distribution of all studies focused on other fields are displayed individually in Fig. 4.  \cite{Yang2024, Ma2024, Lyu2023, Zhang2022, Fang2021}.

\begin{figure}[h!]
\centering
\includegraphics[width=0.8\columnwidth]{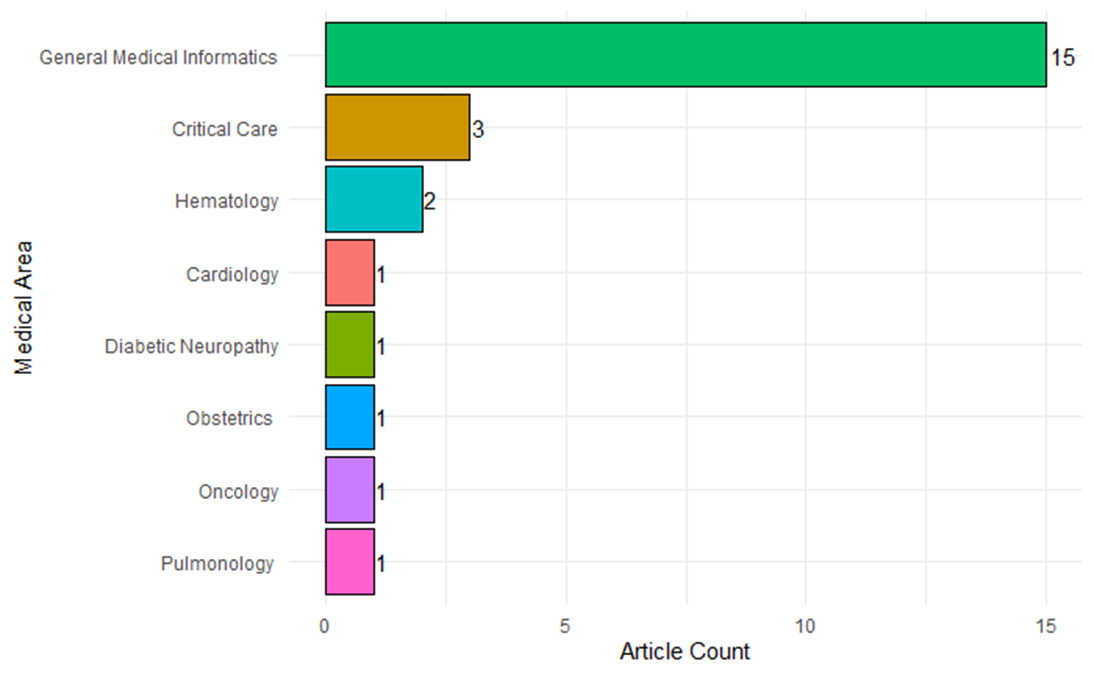} 
\caption{Distribution of the application of GCNs to various medical fields identified by the review.}
\label{fig:figure4}
\end{figure}

\subsection{Hybrid Models}

\par 	Fifteen out of 25 papers surveyed utilized multiple machine learning models to bolster GCN performance. Six papers used a BERT language model as part of their proposed system [26, 27, 31, 32, 37, 39]. Other methods found in multiple studies include RNN-based models and LSTM models [16, 22, 23, 24, 25]. These results show that the use of multiple types of machine learning models is the most common approach within our review.

\subsection{EHR Data Sources}

\par Out of the 25 papers surveyed, eight used real-world datasets [13], [15], [17], [26], [29], [35],  [36], [38]. This is important because it is difficult to get real EHR data and real-world data is often a better indicator of model utility. Among the studies not using real-world data, most used public benchmark EHR datasets. Table 2 shows that the most widely used benchmarks were MIMIC-III and MIMIC IV - used in 10 of the articles surveyed. 

\begin{table*}[h!]
\centering
\renewcommand{\arraystretch}{1.2} 
\resizebox{\textwidth}{!}{%
\begin{tabular}{p{5cm}p{9cm}}
\hline
\textbf{Data Resources} & \textbf{Papers} \\ \hline
MIMIC III & {[}13{]}, {[}15{]}, {[}21{]}, {[}22{]}, {[}23{]}, {[}28{]}, {[}30{]}, {[}33{]}, {[}37{]} \\
MIMIC IV & {[}22{]}, {[}34{]} \\
Other Benchmarks & {[}24{]}, {[}25{]}, {[}27{]}, {[}32{]}, {[}39{]} \\ \hline
Real-world data & {[}13{]}, {[}15{]}, {[}17{]}, {[}26{]}, {[}29{]}, {[}35{]}, {[}36{]}, {[}38{]} \\ \hline
Shanghai Hospital Development Center Clinical Indicator Terminology Base & {[}31{]} \\
Flatiron Health and Foundation Medicine NSCLC clinico-genomic database & {[}40{]} \\
\end{tabular}%
}
\caption{Summary of datasets used in the reviewed articles.}
\label{tab:Table 1}
\end{table*}

\section{Discussion}

\par In this section we will emphasize the key takeaways, provide important context, and discuss recent trends. Potential avenues for further advancement in this domain and the main limitations of this survey conclude the Discussion section.

\subsection{Overview}

\par GCNs are able to represent complex relationships in EHR data and perform convolutions to extract meaningful insights. Using machine learning algorithms on patient data can free up valuable time for medical professionals and improve the quality of healthcare patients receive. Different medical domains often present unique types of problems for machine learning models.  In Fig. 4, we present an overview of which areas of medical science contain the most literature regarding GCNs applied to EHR data. The temporal trend seen in Fig. 3 implies this field has attracted the interest of researchers in 2024.  Hybrid approaches often perform better due to their ability to implement specialized models to perform specific tasks within a prediction pipeline. The most common sources of EHR data are presented in Table 2, with many studies using either publicly available benchmark EHR data or real-world data. 

\subsection{Deep Learning and Tabular Data}

\par Tabular data remains challenging for deep learning architectures. Tree-based models such as XGBoost, Random Forests, and Gradient Boosting Trees remain the most accurate models for tabular data [41]. However, some challenging aspects for deep learning models have been identified by Grinsztajn et al.; they concluded that the irregular patterns, uninformative features, and non-rotationally invariant data are specific characteristics of tabular data that present significant challenges for deep learning models [41]. Consistent with the findings of this survey, other reviews have found most deep neural network approaches for tabular data use a hybrid architecture [42]. Although tree-based models continue to prove more accurate on tabular data, deep learning architectures can be valuable as part of a specialized, hybrid solution.

\subsection{Limitations}

\par The intentionally narrow scope of this paper leaves many important aspects of how GCNs are used on EHR data unexplored. Further analyzing and categorizing the selected articles according to convolution mechanism and type of graph representation would deliver meaningful, pragmatic information. Since we did not report performance metrics, our study is limited in its ability to do any comparative analysis regarding model efficacy. A small number of published papers combining GCNs with EHR data - while emphasizing the need for more research - makes this survey rather brief.

\subsection{Future Work}

\par Deep learning models face several challenges when being used with tabular data. EHRs are typically heterogeneous, tabular data, which are especially challenging for deep learning algorithms. One approach to overcoming this challenge is to convert the tabular data into a pixel representation for subsequent training of convolutional neural networks [43]. Graph Convolution Transformers (GCT) offer a novel approach to leveraging EHR data. Authors in Choi et al. propose a GCT approach that uses known characteristics of the data to guide the transformer in learning the hidden structure in EHRs [44]. These approaches present opportunities to further explore this problem and expand the body of knowledge.

\section{Conclusion}

\par This article surveys several aspects of how GCN models are used with EHR data. The most common medical fields found in this review were medical informatics and critical care. The tasks performed varied widely from natural language processing tasks to clinical risk assessment. Sources of EHR data for model training and evaluation included public benchmark and real-world datasets. The development of better deep learning methods for tabular data is an active area of research that will likely provide parallel advancements for tabular EHR data as well.

\par Here, we have presented a detailed overview of recent research into how researching are using GCNs to leverage EHR data to perform medical tasks. In a field with a dearth of academic literature, this survey illuminates the current state of the domain, in an effort to provide a starting point for future research endeavors. 


\end{document}